\documentclass[10pt, a4paper]{article}

\usepackage[]{lrec-coling2024} 
\usepackage{amsmath}
\usepackage{amsthm}
\usepackage{multirow}
\usepackage{graphicx}
\usepackage{subfigure}
\usepackage{adjustbox}

\title{BootTOD: Bootstrap Task-oriented Dialogue Representations by Aligning Diverse Responses}

\name{Weihao Zeng$^{1}$, Keqing He$^{2}$, Yejie Wang$^{1}$, Dayuan Fu$^{1}$, Weiran Xu$^{1}$ \thanks{Weiran Xu is the corresponding author.}} 

\address{$^1$Beijing University of Posts and Telecommunications, Beijing, China \\
         $^{2}$Meituan, Beijing, China \\
         \texttt{\{zengwh,wangyejie,fdy,xuweiran\}@bupt.edu.cn} \\
         \texttt{
         hekeqing@meituan.com}}

\abstract{
Pre-trained language models have been successful in many scenarios. However, their usefulness in task-oriented dialogues is limited due to the intrinsic linguistic differences between general text and task-oriented dialogues. Current task-oriented dialogue pre-training methods rely on a contrastive framework, which faces challenges such as selecting true positives and hard negatives, as well as lacking diversity. In this paper, we propose a novel dialogue pre-training model called BootTOD. It learns task-oriented dialogue representations via a self-bootstrapping framework. Unlike contrastive counterparts, BootTOD aligns context and context+response representations and dismisses the requirements of contrastive pairs. BootTOD also uses multiple appropriate response targets to model the intrinsic one-to-many diversity of human conversations. Experimental results show that BootTOD outperforms strong TOD baselines on diverse downstream dialogue tasks.
 \\ \newline  }

\begin{document}

\maketitleabstract

\section{Instroduction}


 \begin{figure}[t]
 \centering
\resizebox{0.45\textwidth}{!}{
 \includegraphics[scale=0.7]{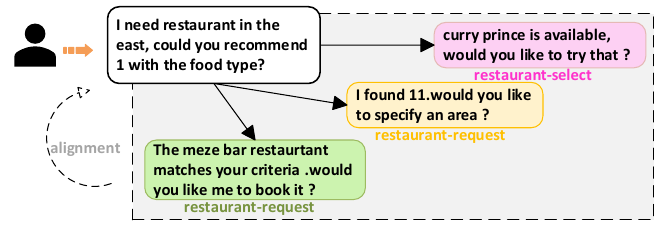}
 }
 \caption{The same context may have multiple appropriate responses in a task-oriented dialogue.}
 \label{fig:intro}
\end{figure}

Previous unsupervised pre-training models for Task-Oriented Dialogues have employed contrastive learning (CL) framework \cite{chen2020simple,He2020MomentumCF}, with the goal of bringing semantically similar (positive) pairs closer together while separating semantically dissimilar (negative) pairs. TOD-BERT \cite{Wu2020TODBERTPN} employs dialogue history and corresponding responses as positive pairs, achieving excellent performance on response selection tasks but only marginal improvements on other dialogue tasks. This is due to the fact that TOD-BERT selects responses from other dialogues as negatives, and these negative responses may be suitable for the current context, resulting in false negatives \cite{Huynh2022BoostingCS,Chen2022IncrementalFN}. Furthermore, DSE \cite{Zhou2022LearningDR} learns from dialogues by using consecutive utterances from the same dialogue as positive pairs. However, this assumption that consecutive utterances represent similar semantics can fail when answers are general and ubiquitous.


 \begin{figure*}[t]
 \centering
\resizebox{0.9\textwidth}{!}{
 \includegraphics[scale=0.5]{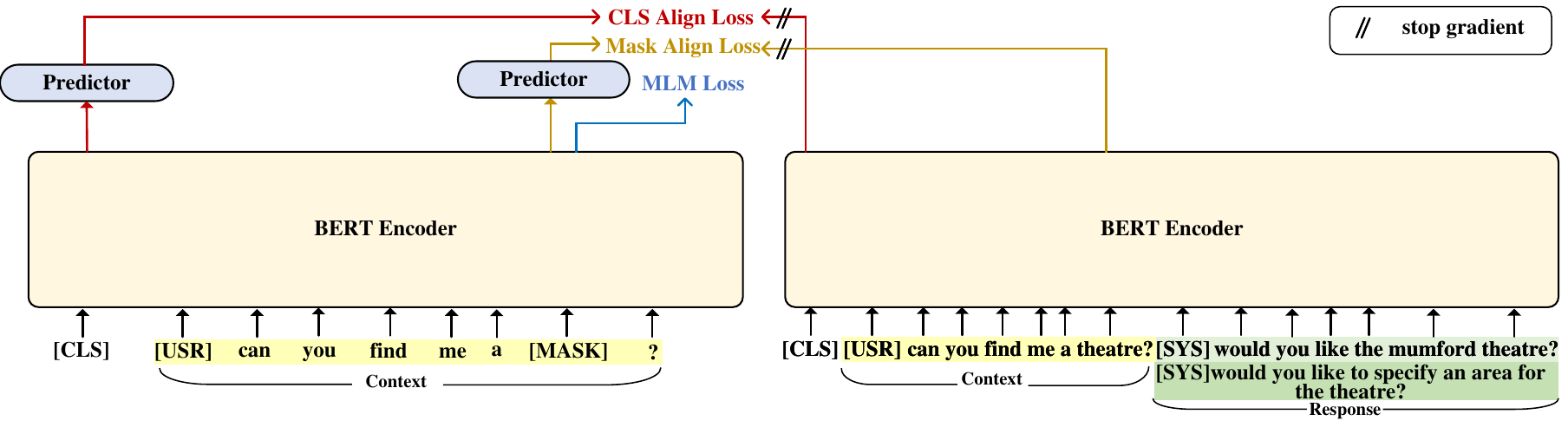}
 }

 \caption{Overall architecture of our proposed BootTOD.}
 \label{fig:model}

\end{figure*}

Despite the remarkable progress of previous TOD PLMs, there are still two challenges. First, these contrastive methods suffer from 
selecting noisy positive and negative pairs, such as false negatives \cite{Huynh2022BoostingCS,Chen2022IncrementalFN}, unreasonable assumptions \cite{Zhou2022LearningDR} and relying on a large batch size \cite{He2020MomentumCF}. Limited exploration has been attempted to perform dialogue pre-training using a non-contrastive framework. Second, most work ignores the one-to-many property in conversation where multiple responses can be appropriate under the same conversation context (as shown in Figure \ref{fig:intro}). PLATO \cite{Bao2019PLATOPD} proposes discrete latent variables to improve utterance-level diversity in open-domain dialog generation, but none of the previous TOD pre-training methods consider such one-to-many property which is also prevalent in task-oriented dialogues. Current TOD PLMs tend to capture the most common dialog policy but ignore rarely occurred yet feasible user behaviors, resulting in duplicate and plain responses. 

To solve the issues, in this paper, we propose a novel dialogue pre-training model, BootTOD, which learns task-oriented dialogue representations via a self-bootstrapping framework. Instead of contrastive counterparts, we introduce a self-bootstrapping framework to align context and context+response representations and dismiss the requirements of contrastive pairs. Besides, BootTOD aligns the context representation with multiple appropriate response targets to model the intrinsic one-to-many diversity of human conversations. Specifically, we use a BERT model to encode the dialogue context and align its representation with the full sequence containing context and response. We argue that a good dialogue representation both learns local context information and predicts future knowledge. Our alignment objectives contain three aspects: dialogue representation alignment using [CLS], [MASK] token representation alignment, and original MLM loss \cite{devlin-etal-2019-bert}. We evaluate BootTOD on various task-oriented dialogue tasks, including intent classification, dialogue state tracking, dialogue act prediction, and response selection. Results show that BootTOD achieves consistent improvements over strong TOD baselines in all the scenarios, which proves its generalization capability. 

Our contributions are: (1) We propose a novel dialogue pre-training model, BootTOD, which uses a self-bootstrapping framework to align the context representation with diverse response targets. (2) Our model outperforms strong TOD baselines on diverse downstream dialogue tasks

\section{Model}

\subsection{Overall Architecture}
Figure \ref{fig:model} displays the overall architecture of BootTOD. Following previous work \cite{Wu2020TODBERTPN,Zhou2022LearningDR, zeng-etal-2023-futuretod}, we adopt BERT-base-uncased\footnote{https://huggingface.co/bert-base-uncased} as our backbone. We add two special role tokens [USR] or [SYS] to the prefix of each utterance and concatenate all the utterances in the same dialogue into one flat sequence. Then we split each dialogue at a randomly selected turn t to get the context and response. We encode the dialogue context via a predictor layer and align its representation with the full sequence containing context and response, including [CLS] alignment, [MASK] token alignment, and mask language model (MLM). We aim to make the model capture local context information and predict future knowledge.

\subsection{Bootstrap Task-oriented Dialogue Representations}

For each dialogue, we first transform it into a token sequence $D=\left\{U_{1}, S_{1}, \ldots, U_{n}, S_{n}\right\}$. $U_{i}$ and $S_{i}$ denote the user utterance and system utterance with a prefix of two special role tokens [USR] or [SYS], respectively. $n$ is the turn number of the dialogue.

Compared to existing contrastive methods, we employ a self-bootstrapping framework to align context and context+response representations to learn future knowledge. The advantages are two-fold: (1) Our framework doesn't require contrastive pairs thus alleviating the noise of selecting positive and negative samples. (2) Learning future knowledge encourages the model to align representations in the same latent space instead of pulling together representations of context and response belonging to different distributions. Assuming we split each dialogue at a randomly selected turn t, the context is $C=\left\{U_{1}, S_{1}, \ldots, U_{t}\right\}$ and the response is $R=\left\{S_{t}, U_{t+1}, S_{t+1}, \ldots, U_{n}, S_{n}\right\}$. Note that in this paper, we denote a response as a multi-turn collection ending with a system utterance. We concatenate all the utterances into sequence and use a shared BERT encoder $f$ to process the context and context+response sequences respectively. Inspired by ~\cite{Chen2020ExploringSS,grill2020bootstrap}, we use a shared predictor MLP head $h$ to transform the representations of the context $C$. We hope the context representation can predict future information while modeling the local semantics. Therefore, we design three alignment objectives as follows.

\textbf{Dialogue Representation Alignment Loss} 
{\setlength{\abovedisplayskip}{0.2cm}
\setlength{\belowdisplayskip}{0.2cm}
\begin{align}
   \mathcal{L}_{cls} = \sum_{l=1}^{L}\left\|h(c_{cls}^{l}) - r_{cls}^{l} \right\|_{2}
\end{align}
\label{cls}
} where $l$ is the $l$-th layer of BERT-base and $h$ is the predictor. $c_{cls}^{l}$ and $r_{cls}^{l}$ are the $l$-th layer [CLS] representations of context and context+response, respectively. We find perform alignment loss on multiple layers rather than only the top layer gives consistent improvements (see Section \ref{layer}). We also try to apply normalization to $c_{cls}^{l},r_{cls}^{l}$ and other forms of objectives but do not observe significant change.

\textbf{Token Representation Alignment Loss} Apart from the dialogue-level alignment, we also propose a token-level alignment loss to learn fine-grained token representations.
{\setlength{\abovedisplayskip}{0.2cm}
\setlength{\belowdisplayskip}{0.2cm}
\begin{align}
   \mathcal{L}_{mask} = \sum_{m=1}^{M}\sum_{l=1}^{L}\left\|h(c_{mask,m}^{l}) - r_{m}^{l} \right\|_{2}
\end{align}
\label{mask}
} where $M$ is the total number of masked tokens. $c_{mask, m}^{l}$ is the $l$-th layer [MASK] token representation of context and $r_{m}^{l}$
is the corresponding original token's $l$-th layer representation of context+response. Note that we only perform mask strategy to the context instead of context+response sequence, which provides more accurate contextual targets to the context representations.

\textbf{Mask Language Model Loss} We also keep the traditional masked language modeling (MLM) \cite{devlin-etal-2019-bert} loss following \citet{Wu2020TODBERTPN}.
{\setlength{\abovedisplayskip}{0.2cm}
\setlength{\belowdisplayskip}{0.2cm}
\begin{align}
   \mathcal{L}_{m l m}=-\sum_{m=1}^M \log P\left(x_m\right)
\end{align}
\label{mlm}
}
where $P(x_{m})$ is the predicted probability of the mask token $x_{m}$ over the vocabulary size.

We simply sum them up and achieve the best performance in our experiments. Inspired by \cite{Chen2020ExploringSS}, we employ a stop-gradient strategy to the representations of context+response as shown in Figure \ref{fig:model} to prevent collapsing. To explore the diversity of different response targets, we randomly select a ratio of consecutive response utterances from $R=\left\{S_{t}, U_{t+1}, S_{t+1}, \ldots, U_{n}, S_{n}\right\}$, such as $\{S_{t}\}$ and $\{S_{t}, U_{t+1}, S_{t+1}\}$. And the last turn of response must be a system utterance. For the same context with multiple appropriate responses, BootTOD aligns the context representation with diverse response targets by iterating over the whole dataset).

\section{Experiment}

\subsection{Training Details}

\textbf{Pre-training Corpus} We utilize nine task-oriented datasets that collected by \citet{Wu2020TODBERTPN}. 


\textbf{Baselines} We compare BootTOD against several strong baselines, including BERT \cite{devlin-etal-2019-bert}, BERT-mlm (continual pre-training on dialogues), DialoGPT \cite{Zhang2020DIALOGPTL}, SimCSE \cite{gao-etal-2021-simcse}, TOD-BERT \cite{Wu2020TODBERTPN}, and DSE \cite{Zhou2022LearningDR}. We focus on unsupervised TOD pre-training so we exclude comparisons with supervised methods that utilize labeled NLI datasets \cite{Williams2018ABC,Welleck2019DialogueNL}  or dialogue act labels\cite{He2022SPACE2TS}.

\begin{table}[t]
\centering
\resizebox{0.48\textwidth}{!}{
\begin{tabular}{c|l|l|l|l|l}
\hline
\multicolumn{1}{l|}{}                                                      & \multicolumn{1}{l|}{\textbf{Model}} & \multicolumn{1}{l|}{\begin{tabular}[c]{@{}l@{}}Acc\\ (all)\end{tabular}} & \multicolumn{1}{l|}{\begin{tabular}[c]{@{}l@{}}Acc\\ (in)\end{tabular}} & \multicolumn{1}{l|}{\begin{tabular}[c]{@{}l@{}}Acc\\ (out)\end{tabular}} & \multicolumn{1}{l}{\begin{tabular}[c]{@{}l@{}}Recall\\ (out)\end{tabular}} \\ \hline
\multirow{6}{*}{\textbf{1-Shot}}                                           & BERT                       & 29.3\%                                                                   & 35.7\%                                                                  & 81.3\%                                                                   & 0.4\%                                                                      \\
                                                                           & BERT-mlm                   & 38.9\%                                                                   & 47.4\%                                                                  & 81.6\%                                                                   & 0.5\%                                                                      \\
                                                                           & SimCSE                     & 29.9\%                                                                   & 36.4\%                                                                  & 81.7\%                                                         & 0.6\%                                                                      \\
                                                                           & TOD-BERT                   & 42.5\%                                                                   & 52.0\%                                                                  & 81.7\%                                                                   & 0.1\%                                                                      \\
                                                                           & DSE                        & 42.3\%                                                                   & 51.7\%                                                                  & \textbf{81.8}\%                                                                   & 0.4\%                                                             \\
                                                                           & BootTOD                  & \textbf{44.0\%}*                                                          & \textbf{53.5\%}*                                                         & 81.7\%                                                                   & \textbf{1.0\%}                                                                     \\ \hline
\multirow{6}{*}{\textbf{10-Shot}}                                                   & BERT                       & 75.5\%                                                                   & 88.6\%                                                                  & 84.7\%                                                                   & 16.5\%                                                                     \\
                                                                           & BERT-mlm                   & 76.6\%                                                                   & 90.5\%                                                                  & 84.3\%                                                                   & 14.0\%                                                                     \\
                                                                           & SimCSE                     & 74.5\%                                                                   & 88.9\%                                                                  & 83.5\%                                                                   & 9.6\%                                                                      \\
                                                                           & TOD-BERT                   & 77.3\%                                                                   & 91.0\%                                                      & 84.5\%                                                                   & 15.3\%                                                                     \\
                                                                           & DSE                        & 77.8\%                                                                   & 90.8\%                                                                  & 85.2\%                                                                   & 19.1\%                                                                     \\
                                                                           & BootTOD                  & \textbf{78.4\%}*                                                          & \textbf{91.1}\%                                                                  & \textbf{85.6\%}*                                                          & \textbf{21.2\%}*                                                            \\ \hline
\multirow{6}{*}{\begin{tabular}[c]{@{}c@{}}\textbf{Full}\\ (\textbf{100-shot})\end{tabular}} & BERT                       & 84.9\%                                                                   & 95.8\%                                                                  & 88.1\%                                                                   & 35.6\%                                                                     \\
& DialoGPT                   & 83.9\%                                                                   & 95.5\%                                                                  & 87.6\%                                                                   & 32.1\%                                                                     \\
                                                                           & BERT-mlm                   & 85.9\%                                                                   & 96.1\%                                                                  & 89.5\%                                                                   & 46.3\%                                                                     \\
                                                                           & SimCSE                     & 82.3\%                                                                   & 94.7\%                                                                  & 86.6\%                                                                   & 26.6\%                                                                     \\
                                                                           & TOD-BERT                   & 86.6\%                                                                   & \textbf{96.2\%}                                                         & 89.9\%                                                                   & 43.6\%                                                                     \\
                                                                           & DSE                        & 84.3\%                                                                   & 95.8\%                                                                  & 87.7\%                                                                   & 32.5\%                                                                     \\
                                                                           & BootTOD                  & \textbf{88.2\%}*                                                          & 96.1\%                                                                  & \textbf{91.1\%}*                                                          & \textbf{52.7\%}*                                                            \\ \hline
\end{tabular}
}

\caption{Intent recognition results on the OOS dataset. Acc(all), Acc(in), Acc(out) denotes the overall accuracy, in-domain intent accuracy and out-of-domain intent accuracy. The numbers with * are significant using t-test with $p < 0.01$.}
\label{main_intent}

\end{table}

\begin{table}[t]
\centering
\resizebox{0.5\textwidth}{!}{
\begin{tabular}{l|ll|ll|ll}
\hline
\multicolumn{1}{c|}{}                                 & \multicolumn{2}{c|}{\textbf{5 \% Data}} & \multicolumn{2}{c|}{\textbf{10 \% Data}}                                          & \multicolumn{2}{c}{\textbf{Full Data}}                    \\
\multicolumn{1}{c|}{\multirow{-2}{*}{\textbf{Model}}} & \textbf{Joint Acc}  & \textbf{Slot Acc} & \textbf{Joint Acc}                      & \textbf{Slot Acc}                       & \textbf{Joint Acc}                      & \textbf{Slot Acc}                       \\ \hline
BERT                                                  & 19.6\%              & 92.0\%            & 32.9\%                                  & 94.7\%                                  & 45.6\%          & 96.6\%          \\
BERT-mlm                                              & 28.1\%              & 93.9\%            & 39.5\%          & 95.6\%          & 47.7\%                                  & 96.8\%                                  \\
SimCSE                        & 21.1\%              & 91.6\%            & 35.6\%                                  & 95.0\%                                  & 48.0\%          & 96.8\%          \\
TOD-BERT                                              & 28.6\%              & 93.8\%            & 37.0\%                                  & 95.2\%                                  & 48.0\%          & 96.9\%          \\
DSE                                                   & 23.8\%              & 93.0\%            & 37.8\%                                  & 95.5\%                                  & 49.9\%          & 97.0\%          \\
BootTOD                                           & \textbf{30.3\%}*     & \textbf{94.2\%}*   & \textbf{40.8\%}* & \textbf{96.0\%}* & \textbf{50.7\%}* & \textbf{97.2\%} \\ \hline
\end{tabular}
}

\caption{Dialogue state tracking results on MWOZ 2.1. Joint Acc and Slot Acc denote the joint goal accuracy and slot accuracy. The numbers with * are significant using t-test with $p < 0.01$.}
\label{main_dst}

\end{table}

\begin{table}[t]
\centering
\resizebox{0.50\textwidth}{!}{
\begin{tabular}{c|l|ll|ll}
\hline
\multirow{2}{*}{}                   & \multicolumn{1}{c|}{\multirow{2}{*}{\textbf{Model}}} & \multicolumn{2}{c|}{\textbf{MWOZ}}                 & \multicolumn{2}{c}{\textbf{DSTC2}}                  \\ \cline{3-6} 
                                    & \multicolumn{1}{c|}{}                                & \multicolumn{1}{c|}{micro-F1}        & macro-F1        & \multicolumn{1}{c|}{micro-F1}        & macro-F1        \\ \hline
\multirow{6}{*}{\textbf{1\% Data}}  & BERT                                                 & \multicolumn{1}{l|}{84.0\%}          & 66.7\%          & \multicolumn{1}{l|}{77.1\%}          & 25.8\%          \\
                                    & BERT-mlm                                             & \multicolumn{1}{l|}{87.5\%}          & 73.3\%          & \multicolumn{1}{l|}{79.6\%}          & 26.4\%          \\
                                    & SimCSE                                               & \multicolumn{1}{l|}{81.0\%}          & 62.1\%          & \multicolumn{1}{l|}{78.9\%}          & 27.3\%          \\

                                    & TOD-BERT                                         & \multicolumn{1}{l|}{86.9\%}          & 72.4\%          & \multicolumn{1}{l|}{82.9\%}          & 28.0\%          \\
                                    & DSE                                                  & \multicolumn{1}{l|}{82.9\%}          & 65.1\%          & \multicolumn{1}{l|}{72.4\%}          & 21.4\%          \\
                                    & BootTOD                                            & \multicolumn{1}{l|}{\textbf{87.7\%}} & \textbf{73.8\%}* & \multicolumn{1}{l|}{\textbf{85.8\%}*} & \textbf{33.9\%}* \\ \hline
\multirow{6}{*}{\textbf{10\% Data}} & BERT                                                 & \multicolumn{1}{l|}{89.7\%}          & 78.4\%          & \multicolumn{1}{l|}{88.2\%}          & 34.8\%          \\
                                    & BERT-mlm                                             & \multicolumn{1}{l|}{90.1\%}          & 78.9\%          & \multicolumn{1}{l|}{91.8\%}          & 39.4\%          \\
                                    & SimCSE                                               & \multicolumn{1}{l|}{89.6\%}          & 77.8\%          & \multicolumn{1}{l|}{92.3\%}          & 40.5\%          \\
                                    & TOD-BERT                                         & \multicolumn{1}{l|}{90.2\%}          & 79.6\%          & \multicolumn{1}{l|}{90.6\%}          & 38.8\%          \\
                                    & DSE                                                  & \multicolumn{1}{l|}{89.9\%}          & 79.4\%          & \multicolumn{1}{l|}{91.1\%}          & 39.0\%          \\
                                    & BootTOD                                            & \multicolumn{1}{l|}{\textbf{90.9\%}*} & \textbf{80.7\%}* & \multicolumn{1}{l|}{\textbf{93.9\%}*} & \textbf{42.8\%} \\ \hline
\multirow{7}{*}{\textbf{Full Data}} 
                                    & BERT                                                 & \multicolumn{1}{l|}{91.4\%}          & 79.7\%          & \multicolumn{1}{l|}{92.3\%}          & 40.1\%          \\

& DialoGPT                                             & \multicolumn{1}{l|}{91.2\%}          & 79.7\%          & \multicolumn{1}{l|}{93.8\%}          & 42.1\%          \\
                                    & BERT-mlm                                             & \multicolumn{1}{l|}{91.7\%}          & 79.9\%          & \multicolumn{1}{l|}{90.9\%}          & 39.9\%          \\
                                    & SimCSE                                               & \multicolumn{1}{l|}{91.6\%}          & 80.3\%          & \multicolumn{1}{l|}{91.5\%}          & 39.6\%          \\

                                    & TOD-BERT                                         & \multicolumn{1}{l|}{91.7\%}          & 80.6\%          & \multicolumn{1}{l|}{93.8\%}          & 41.3\%          \\
                                    & DSE                                                  & \multicolumn{1}{l|}{91.7\%}          & 81.3\%          & \multicolumn{1}{l|}{92.6\%}          & 40.2\%          \\
                                    & BootTOD                                            & \multicolumn{1}{l|}{\textbf{91.8\%}} & \textbf{82.3\%}* & \multicolumn{1}{l|}{\textbf{95.9\%}*} & \textbf{46.5\%}* \\ \hline
\end{tabular}
}

\caption{Dialogue act prediction results on MWOZ and DSTC2. The numbers with * are significant using t-test with $p < 0.01$.}
\label{main_act}

\end{table}

\textbf{Pre-trainging Details} BootTOD's training uses a batch size of 48, a maximum input length of 512, and initiates with BERT-base-uncased. It's optimized with Adam, a learning rate of 5e-5, and 0.2 dropout. The mask ratio is 15\%, and the predictor head has two layers plus ReLU, with dimensions of 768 and 512. After pre-training, we retain the Bert encoder parameters and remove the MLP head for subsequent fine-tuning. The 3-day pre-training involves an early-stop strategy based on perplexity, using eight NVIDIA Tesla A100 GPUs.

\textbf{Finetuning Details} For BERT-mlm and TOD-BERT, we directly use the results reported by TOD-BERT \cite{Wu2020TODBERTPN}. We adopt the same hyperparameters for all downstream tasks.

\subsection{Main Results}


We evaluated various pre-trained language models on four core task-oriented dialogue tasks (We detail the evaluation tasks and evaluation metrics in the Appendix \ref{sec:task_detail}.). We conducted experiments using the whole dataset, as well as a few-shot setting. The few-shot setting here aligns with TOD-BERT \cite{Wu2020TODBERTPN} and FutureTOD \cite{zeng-etal-2023-futuretod}. Specifically, this involves fine-tuning using just 1\% or 10\% of the entire dataset, as opposed to using the full dataset for fine-tuning. The few-shot experiments were randomly sampled at least three times with different seeds. 



\textbf{Intent Recognition} Table \ref{main_intent} displays the results of intent recognition on the OOS dataset \cite{larson-etal-2019-evaluation}. We observe that BootTOD outperforms all the baselines on 10 of 12 metrics, particularly with significant improvements in overall accuracy and OOD metrics. These results demonstrate the generalization ability of BootTOD across both in-domain and out-of-domain metrics.


\textbf{Dialogue State Tracking} Table \ref{main_dst} shows the results of dialogue state tracking on MWOZ 2.1. Our BootTOD achieves state-of-the-art results on all the metrics.  We find SimCSE performs poorly in the 5\% data setting because it ignores the intrinsic properties of dialogue data and can not model overall dialogue well with few data. Our method achieves a greater improvement on joint accuracy than on slot accuracy, indicating the strength of understanding the overall dialogue context. We also find that these baselines overfit to the easy slot values, but can't predict the hard ones, resulting in comparable slot accuracy but poor joint accuracy. For example, BootTOD outperforms TOD-BERT by 0.3\% on Slot Acc but 2.7\% on Joint Acc in the full data setting, which indicates the superiority of dialogue modeling.

\begin{table}[t]
\centering
\resizebox{0.50\textwidth}{!}{
\begin{tabular}{c|l|ll|ll}
\hline
                                     & \multicolumn{1}{c|}{}                                 & \multicolumn{2}{c|}{\textbf{MWOZ}}                                                                 & \multicolumn{2}{c}{\textbf{DSTC2}}                                                \\ \cline{3-6} 
\multirow{-2}{*}{}                   & \multicolumn{1}{c|}{\multirow{-2}{*}{\textbf{Model}}} & \multicolumn{1}{c|}{1-to-100}                                & 3-to-100                                & \multicolumn{1}{c|}{1-to-100}                       & 3-to-100                       \\ \hline
                                     & BERT                                                  & \multicolumn{1}{l|}{7.8\%}                                   & 20.5\%                                  & \multicolumn{1}{l|}{3.7\%}                          & 9.6\%                          \\
                                     & BERT-mlm                                              & \multicolumn{1}{l|}{13.0\%}                                  & 34.6\%                                  & \multicolumn{1}{l|}{12.5\%}                         & 24.9\%                         \\
                                     & SimCSE                                                & \multicolumn{1}{l|}{17.2\%}                                  & 32.6\%                                  & \multicolumn{1}{l|}{27.6\%}                         & 46.4\%                         \\
                                     & TOD-BERT                                          & \multicolumn{1}{l|}{-}               & -               & \multicolumn{1}{l|}{37.5\%} & 55.9\% \\
                                     & DSE                                                   & \multicolumn{1}{l|}{7.9\%}                                   & 21.2\%                                  & \multicolumn{1}{l|}{2.4\%}                          & 6.1\%                          \\
\multirow{-6}{*}{\textbf{1\% Data}}  & BootTOD                                             & \multicolumn{1}{l|}{\textbf{37.0\%}*} & \textbf{60.5\%}* & \multicolumn{1}{l|}{\textbf{38.1\%}*}                & \textbf{61.3\%}*                \\ \hline
                                     & BERT                                                  & \multicolumn{1}{l|}{20.9\%}                                  & 45.4\%                                  & \multicolumn{1}{l|}{8.9\%}                          & 21.4\%                         \\
                                                                          & BERT-mlm                                              & \multicolumn{1}{l|}{22.3\%}          & 48.7\%          & \multicolumn{1}{l|}{19.0\%} & 33.8\% \\
                                     & SimCSE                                                & \multicolumn{1}{l|}{37.2\%}          & 60.6\%          & \multicolumn{1}{l|}{42.0\%} & 63.5\% \\

                                     & TOD-BERT                                          & \multicolumn{1}{l|}{-}               & -               & \multicolumn{1}{l|}{49.7\%}                         & 66.6\%                         \\
                                     & DSE                                                   & \multicolumn{1}{l|}{24.8\%}          & 49.4\%          & \multicolumn{1}{l|}{42.0\%} & 59.7\% \\
\multirow{-6}{*}{\textbf{10\% Data}} & BootTOD                                             & \multicolumn{1}{l|}{\textbf{50.0\%}*}                         & \textbf{72.0\%}*                         & \multicolumn{1}{l|}{\textbf{52.3\%}*}                & \textbf{69.6\%}*                \\ \hline
                                     & BERT                                                  & \multicolumn{1}{l|}{47.5\%}          & 75.5\%          & \multicolumn{1}{l|}{46.6\%} & 62.1\% \\
& DialoGPT                                              & \multicolumn{1}{l|}{35.7\%}                                  & 64.1\%                                  & \multicolumn{1}{l|}{39.8\%}                         & 57.1\%                         \\
                                     & BERT-mlm                                              & \multicolumn{1}{l|}{48.1\%}                                  & 74.3\%                                  & \multicolumn{1}{l|}{50.0\%}                         & 65.1\%                         \\
                                     & SimCSE                                                & \multicolumn{1}{l|}{64.2\%}                                  & 85.4\%                                  & \multicolumn{1}{l|}{55.6\%}                         & 70.5\%                         \\
                                     & TOD-BERT                                          & \multicolumn{1}{l|}{65.8\%}          & 87.0\%          & \multicolumn{1}{l|}{56.8\%} & 70.6\% \\
                                     & DSE                                                   & \multicolumn{1}{l|}{63.3\%}                                  & 85.3\%                                  & \multicolumn{1}{l|}{58.3\%}                         & 72.0\%                         \\
\multirow{-7}{*}{\textbf{Full Data}} & BootTOD                                             & \multicolumn{1}{l|}{\textbf{68.8\%}*}                         & \textbf{87.6\%}*                         & \multicolumn{1}{l|}{\textbf{59.1\%}*}                & \textbf{72.3\%}                \\ \hline
\end{tabular}
}

\caption{Response selection results on MWOZ and DSTC2. 1-to-100 and 3-to-100 denote the ratio of the ground-truth response being ranked at the top-1 or top-3 given 100 candidates. The numbers with * are significant using t-test with $p < 0.01$.}
\label{main_rs}

\end{table}

\textbf{Dialogue Act Prediction} Table \ref{main_act} shows the results of dialogue act prediction on MWOZ and DSTC2. Our BootTOD achieves state-of-the-art results on all the metrics. We find our method obtains comparable performance only using 10\% data than the baselines using 100\% data, which verifies the superior few-shot learning capability.


\textbf{Response Selection} Table \ref{main_rs} displays the results of response selection on MWOZ and DSTC2.\footnote{TOD-BERT uses the response contrastive loss as the pre-training objective on full MWOZ training data so we don't report its results on few-shot setting.} Our BootTOD achieves state-of-the-art results on all the metrics. We find DSE performs poorly in the 1\% data setting and even worse than BERT on DSTC2.  It shows the assumption that consecutive utterances represent similar semantics fails in practical dialogue scenarios. Although TOD-BERT is pre-trained with a response contrastive objective, our method still outperforms it on DSTC2 significantly both in full data and few data settings.  It indicates that BootTOD can achieve better generalization capability.

\begin{table}[t]
\centering
\resizebox{0.5\textwidth}{!}{
\begin{tabular}{l|cc|cc}
\hline
\multicolumn{1}{c|}{\multirow{2}{*}{Model}} & \multicolumn{2}{c|}{\textbf{DSTC2}}                                            & \multicolumn{2}{c}{\textbf{MWOZ}}                                             \\
\multicolumn{1}{c|}{}                       & \multicolumn{1}{l}{micro-F1} & \multicolumn{1}{l|}{macro-F1} & \multicolumn{1}{l}{1-to-100} & \multicolumn{1}{l}{3-to-100} \\ \hline
BootTOD                                 & \textbf{95.85\%}                      & \textbf{46.53\%}                       & 68.79\%                               & 87.61\%                               \\
w/o Mask Align                            & 95.58\%                               & 46.17\%                                & 68.74\%                               & 87.70\%                               \\
w/o CLS Align                             & 95.06\%                               & 45.37\%                                & 67.11\%                               & 87.38\%                               \\
w/o Stop Gradient                           & 95.50\%                               & 46.13\%                                & \textbf{68.86\%}                      & \textbf{88.16\%}                      \\
w/o MLP Head                                & 95.03\%                               & 45.65\%                                & 68.34\%                               & 87.67\%                               \\ \hline
\end{tabular}
}

\caption{Ablation study Results. Removing the MLM will make BootTOD fail to converge, so we do not report this result.}
\label{abla_result}

\end{table}



\section{Qualitative Analysis}

\subsection{Ablation Study}

Table \ref{abla_result} presents the ablation results of dialogue act prediction on DSTC2 and response selection on MWOZ. BootTOD without CLS Align performs the worst among all the variations. This indicates that CLS alignment loss is crucial for capturing the dialogue-level information, allowing the dialogue model to have better representation capabilities. Removing MLP Head also damages the performance. We find that 
removing MLP head makes the training unstable and adding a predictor serves as a decoder to learn future representation. 
Mask Align also contributes to performance, illustrating the importance of learning fine-grained token representations. Besides, the Stop gradient has a positive impact on dialogue act prediction but a negative impact on response selection. We believe it is due to the mismatch between the stop-gradient and the dual-encoder used in the response selection task.

\subsection{Hyper-parameter Analysis}

\label{layer}
\begin{figure}[t]
    \centering
    \begin{adjustbox}{minipage=\linewidth,scale=1.0}
    \subfigure[micro-F1]{
        \includegraphics[width=0.48\textwidth]{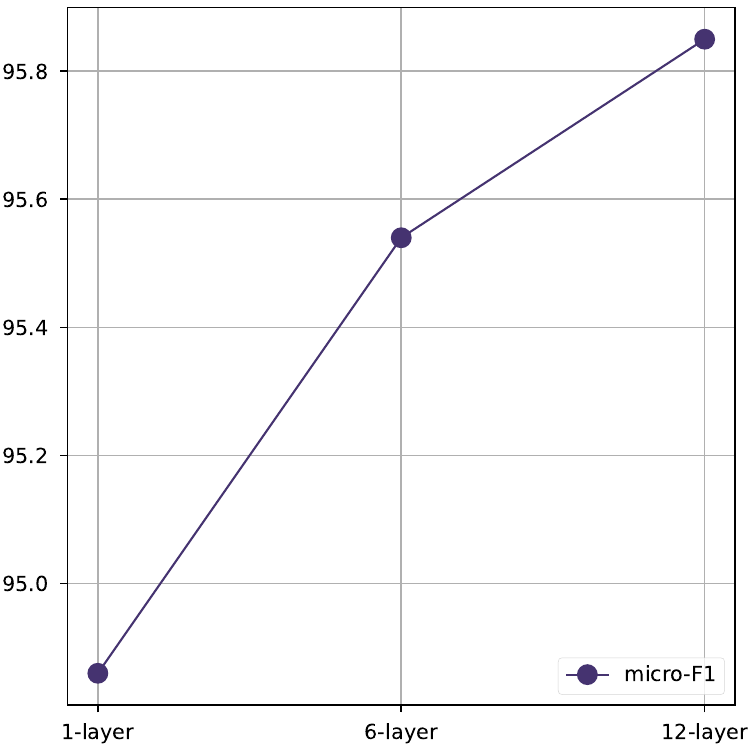}
    }
    \hspace{-0.3cm}
    \subfigure[macro-F1]{
        \includegraphics[width=0.48\textwidth]{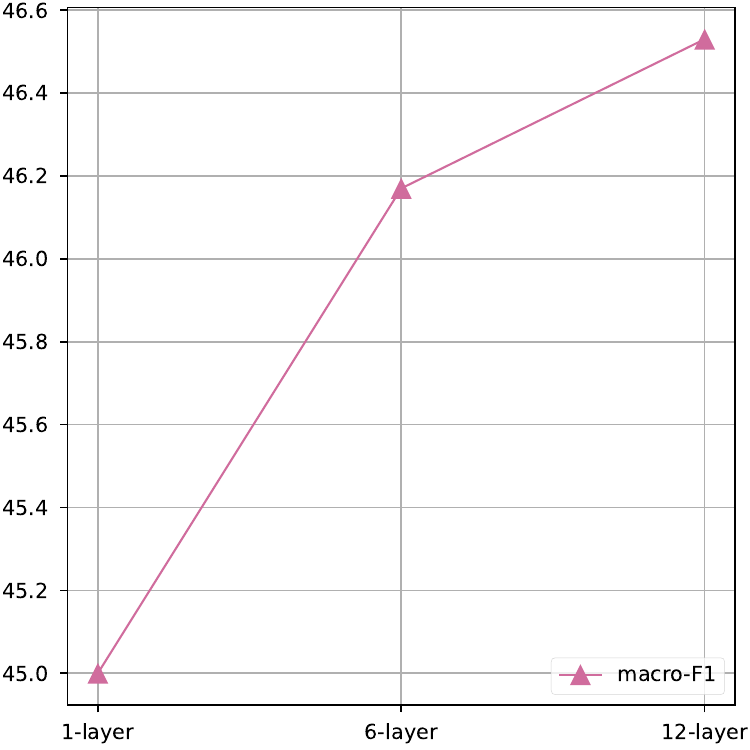}
    }
    \caption{Ablation study of Alignment Layers. We report the results of dialogue act prediction on DSTC2. The X-asix and Y-asix denotes the number of layers used for alignment and performance.}
    \label{fig:top_layer}
    \end{adjustbox}
\end{figure}

\textbf{Effect of Alignment Layers}
BootTOD uses the top-K Layer Representation for alignment loss $L_{cls}$ and $L_{mask}$. Figure \ref{fig:top_layer} shows the effect of varying the value of K. We find BootTOD gets improvements as the value of K increases.  It indicates that different layers of the model can capture features of different granularities, thereby improving the performance of the downstream tasks.

\begin{figure}[t]
    \centering
    \begin{adjustbox}{minipage=\linewidth,scale=1.0}
    \subfigure[micro-F1]{
        \includegraphics[width=0.48\textwidth]{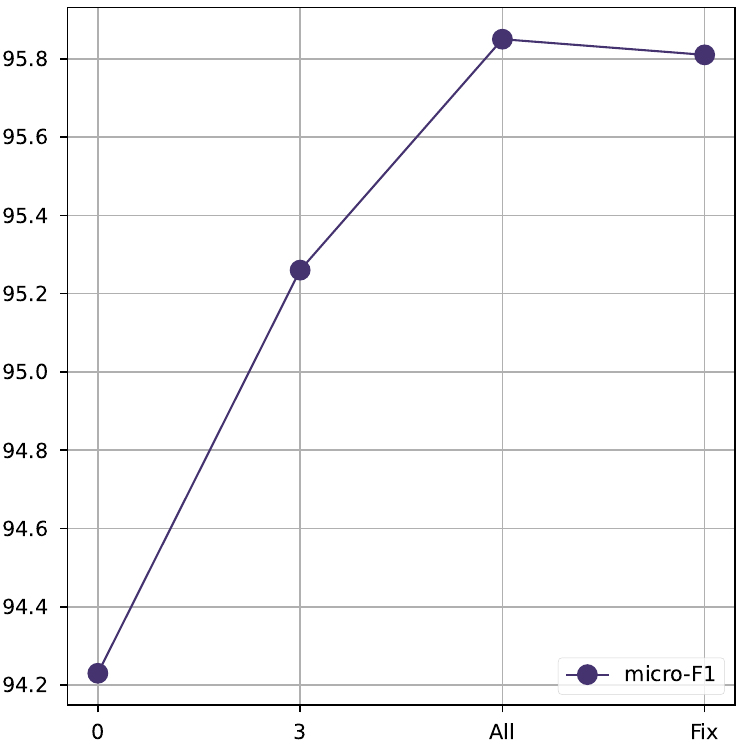}
    }
    \hspace{-0.3cm}
    \subfigure[macro-F1]{
        \includegraphics[width=0.48\textwidth]{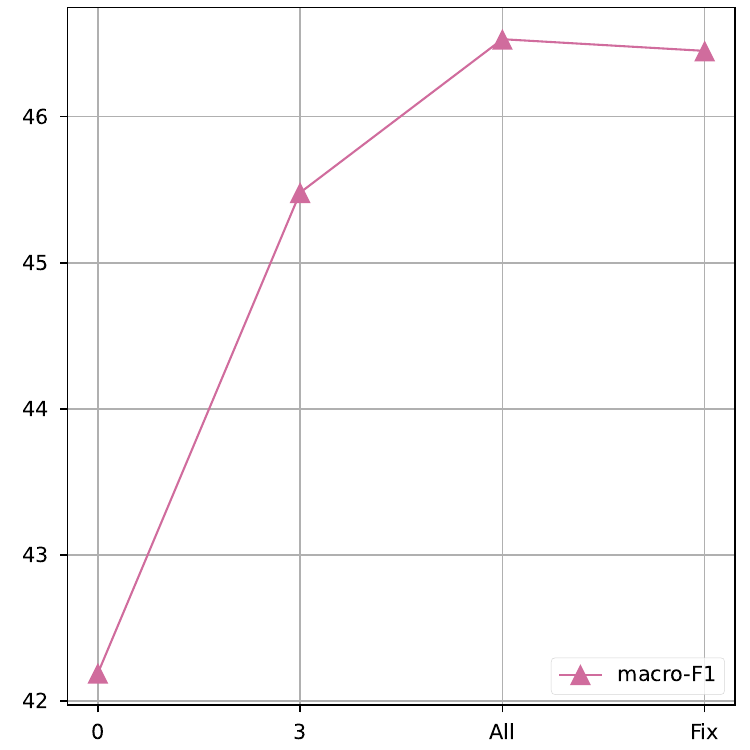}
    }
    \caption{Ablation study of max future length $P$. We report the results of dialogue act prediction on DSTC2. The X-asix and Y-asix denotes the max future length $P$ and performance.}
    \label{fig:fut_len}

    \end{adjustbox}
\end{figure}

\textbf{Effect of Max Response Length} The response consists of consecutive utterances, and we set the number of selectable utterances from 1 to max response length $P$. To explore the effect of varying the value of $P$, we set the $P$ to 0, 3, $All$, and $Fix$ respectively.  $P=All$ denotes that we can randomly select any length of utterances from the whole utterances, while $P=Fix$ denotes that we must use the whole consecutive future utterances together. For example, if we have 5 future utterances $F=\{S_{t}, U_{t+1}, S_{t+1}, U_{t+2}, S_{t+2}\}$. $P=3$ allows us to select any length no longer than 3, such as  $\{S_{t}\}$ or $\{S_{t}, U_{t+1}, S_{t+1}\}$; $P=All$ allows us to select any length of future from the 5 utterances, that is $\{S_{t}\}$ or $\{S_{t}, U_{t+1}, S_{t+1}\}$ or F; $P=Fix$ can only select $F$. Figure \ref{fig:fut_len} shows BootTOD generally gets improvements with increasing the $P$, indicating that more response targets are beneficial to learn more diverse dialogue representations. We also find that $P=Fix$ degrades performance compared to $P=All$. We argue that fixed response information will narrow down dialogue context representation space. 

\section{Non-Contrastive Methods Comparison}

As supervised methods rely on labeled NLI datasets \cite{Williams2018ABC,Welleck2019DialogueNL} or dialogue act labels \cite{He2022SPACE2TS}, we didn't include them in a fairness comparison. Instead, we compared BootTOD with a recent non-contrastive method, FutureTOD \cite{zeng-etal-2023-futuretod}. FutureTOD proposes a non-contrastive framework that distills future knowledge into the representation of the previous dialogue. The results are displayed in Table \ref{non_con_ir}, Table \ref{non_con_dst}, Table \ref{non_con_dap}, and Table \ref{non_con_rs} in the Appendix. Our method has demonstrated excellent performance on most metrics across all tasks compared to FutureTOD. This underscores the improvement of our BootTOD's performance in comparison to other non-contrastive methods.

\section{Conclusion}

In this paper, we propose a novel dialogue pre-training model, BootTOD, which learns task-oriented dialogue representations via a self-bootstrapping framework. Instead of contrastive counterparts, BootTOD aligns context and context+response representations and dismisses the requirements of contrastive pairs. Besides, BootTOD aligns the context representation with diverse targets to model the intrinsic one-to-many diversity of human conversations. We perform comprehensive experiments on various task-oriented dialogue tasks. BootTOD significantly outperforms TOD-BERT, DSE, and other strong baselines in all the scenarios. 

\section*{Acknowledgements}
We thank all anonymous reviewers for their helpful comments and suggestions. This work was supported by the National Natural Science Foundation of China (NSFC No.62076031 and No.62076036). This work is partially supported by State Key Laboratory of Massive Personalized Customization System and Technology (No. H\&C-MPC-2023-02-07(Q)).

\nocite{*}
\section{Bibliographical References}\label{sec:reference}

\bibliographystyle{lrec-coling2024-natbib}
\bibliography{lrec-coling2024-example}


\appendix

\section{Evaluation Details}

\label{sec:task_detail}
We evaluated various pre-trained language models on four core task-oriented dialogue tasks, including intent recognition, dialogue state tracking, dialogue act prediction, and response selection. Here, we provide more details about these evaluation tasks and metrics.

\noindent\textbf{Intent Recognition} is a multi-class classification task that takes a dialogue utterance as input and predicts an intent label \cite{zeng-etal-2022-semi}. We use the [CLS] embeddings from the model as the dialogue representation. The model is trained with cross-entropy loss. We report classification accuracy and recall.

\noindent\textbf{Dialogue State Tracking} is a multi-class classification task, which involves identifying the slot values for each (domain, slot) pair at each dialogue turn, based on a pre-defined ontology. The model takes dialogue history as input and is trained with cross-entropy loss summed over all the pairs. We use a widely-used TOD dataset MWOZ 2.1\cite{Budzianowski2018MultiWOZA} across seven different domains. We report the Joint acc and Slot acc. The Joint acc considers true if and only if the predicted values exactly match its ground truth values at each dialogue turn. The slot acc individually compares each (domain, slot, value) triplet to its ground truth label.

\noindent\textbf{Dialogue Act Prediction} is a multi-label classification task that takes dialogue history as input and predicts multiple dialogue acts corresponding to system response. The model is trained with binary cross-entropy loss over all possible actions. During inference, the threshold for triggering the dialogue act is set to 0.5. We use two datasets MWOZ \cite{Budzianowski2018MultiWOZA} and DSTC2 \cite{Henderson2014TheSD}. Following TODBERT \cite{Wu2020TODBERTPN}, we use the same data preprocessing to uniform the original dialogue acts to a general format. We report the micro-F1 and macro-F1.

\noindent\textbf{Response Selection} is a ranking task that aims to retrieve the most relative system response from a candidate pool based on dialogue history. We also use MWOZ and DSTC2 as our evaluation datasets. We use a dual-encoder strategy, which calculates similarity scores between dialogue history and candidate responses. We train this model with random system responses from the corpus as negative samples. We report k-to-100 accuracy. This metric represents the ratio of the ground-truth response being ranked in the top-k positions when compared to 99 randomly sampled responses, as determined by the scores computed by the dual-encoder.

\section{Non-Contrastive Methods Comparison}

We present the performance of non-contrastive methods in intent recognition, dialogue state tracking, dialogue act prediction, and response selection in the Table \ref{non_con_ir}, Table \ref{non_con_dst}, Table \ref{non_con_dap} and Table \ref{non_con_rs} respectively.

\begin{table}[t]
\centering
\resizebox{0.5\textwidth}{!}{
\begin{tabular}{lcccc}
\hline
          & \textbf{Acc(all)} & \textbf{Acc(in)} & \textbf{Acc(out)} & \textbf{Recall(out)} \\ \hline
FutureTOD & 87.2\%            & 96.0\%           & 90.0\%            & 47.6\%               \\
BootTOD   & \textbf{88.2\%}   & \textbf{96.1\%}  & \textbf{91.1\%}   & \textbf{52.7\%}      \\ \hline
\end{tabular}
}

\caption{The performance of non-contrastive methods on the OOS dataset for Intent recognition. Acc(all), Acc(in), Acc(out) denotes the overall accuracy, in-domain intent accuracy, and out-of-domain intent accuracy.}
\label{non_con_ir}

\end{table}

\begin{table}[t]
\centering
\resizebox{0.35\textwidth}{!}{
\begin{tabular}{lcc}
\hline
\textbf{} & \textbf{Joint Acc} & \textbf{Slot Acc} \\ \hline
FutureTOD & 50.4\%             & 97.1\%            \\
BootTOD   & \textbf{50.7}\%             & \textbf{97.2}\%            \\ \hline
\end{tabular}
}

\caption{The performance of non-contrastive methods on the MWOZ 2.1 for Dialogue State Tracking. Joint Acc and Slot Acc denote the joint goal accuracy and slot accuracy.}
\label{non_con_dst}

\end{table}

\begin{table}[t]
\centering
\resizebox{0.5\textwidth}{!}{
\begin{tabular}{lcccc}
\hline
          & \multicolumn{2}{c}{\textbf{MWOZ}} & \multicolumn{2}{c}{\textbf{DSTC2}} \\
          & \textbf{micro-F1}    & \textbf{macro-F1}   & \textbf{micro-F1}    & \textbf{macro-F1}    \\ \hline
FutureTOD & \textbf{92.0}\%      & 81.9\%     & 94.6\%      & 44.6\%      \\
BootTOD   & 91.8\%      & \textbf{82.3}\%     & \textbf{95.9}\%      & \textbf{46.5}\%      \\ \hline
\end{tabular}
}

\caption{The performance of non-contrastive methods on the MWOZ and DSTC2 for Dialogue Act Prediction.}
\label{non_con_dap}

\end{table}

\begin{table}[t]
\centering
\resizebox{0.5\textwidth}{!}{
\begin{tabular}{lcccc}
\hline
          & \multicolumn{2}{c}{\textbf{MWOZ}}     & \multicolumn{2}{c}{\textbf{DSTC2}}    \\
          & \textbf{1-to-100} & \textbf{3-to-100} & \textbf{1-to-100} & \textbf{3-to-100} \\ \hline
FutureTOD & 68.5\%            & 87.9\%            & 58.4\%            & 72.6\%            \\
BootTOD   & \textbf{68.8}\%            & 87.6\%            & \textbf{59.1}\%            & 72.3\%            \\ \hline
\end{tabular}
}

\caption{The performance of non-contrastive methods on the MWOZ and DSTC2 for Response Selection. 1-to-100 and 3-to-100 denote the ratio of the ground-truth response being ranked at the top-1 or top-3 given 100 candidates. }
\label{non_con_rs}

\end{table}

\end{document}